\documentclass{article}
\usepackage{spconf,amsmath,graphicx}
\usepackage{textcomp}
\usepackage{amssymb}
\usepackage{subfig}

\title{Semi-supervised Cervical Dysplasia Classification with Learnable Graph Convolutional Network}
\name{\resizebox{0.98\textwidth}{!}{Yanglan Ou$^{1}$, Yuan Xue$^{1}$, Ye Yuan$^{2}$, Tao Xu$^{3}$, Vincent Pisztora$^{1}$, Jia Li$^{1}$, Xiaolei Huang$^{1}$}}
\address{$^1$The Pennsylvania State University, University Park\ \ \ \ $^2$Carnegie Mellon University\ \ \ \ $^3$Facebook}
\begin{document}
\maketitle
\begin{abstract}
    Cervical cancer is the second most prevalent cancer affecting women today. As the early detection of cervical carcinoma relies heavily upon screening and pre-clinical testing, digital cervicography has great potential as a primary or auxiliary screening tool, especially in low-resource regions due to its low cost and easy access.
    Although an automated cervical dysplasia detection system has been desirable, traditional fully-supervised training of such systems requires large amounts of annotated data which are often labor-intensive to collect. To alleviate the need for much manual annotation, we propose a novel graph convolutional network (GCN) based semi-supervised classification model
    that can be trained with fewer annotations. In existing GCNs, graphs are constructed with fixed features and can not be updated during the learning process. This limits their ability to exploit new features learned during graph convolution. In this paper, we propose a novel and more flexible GCN model with a feature encoder that adaptively updates the adjacency matrix during learning and demonstrate that this model design leads to improved performance. 
    Our experimental results on a cervical dysplasia classification dataset show that the proposed framework outperforms previous methods under a semi-supervised setting, especially when the labeled samples are scarce.
\end{abstract}

\vspace{-8pt}
\section{Introduction}
\vspace{-6pt}
\label{sec:intro}
Cervical cancer is the second most common type of cancer affecting women globally~\cite{berman2018cervical}. The abnormal growth (potentially precancerous transformation) of cells on the surface of the cervix is known as cervical intraepithelial neoplasia (CIN) or cervical dysplasia, which can be divided into three grades: CIN1, CIN2, and CIN3. 
CIN1 represents mild dysplasia that will usually be cleared by an immune response within one year. 
CIN2 and CIN3 indicate moderate and severe lesions, respectively. 
While dysplasia in CIN1 only needs conservative observation, lesions in CIN2/3 and cancer (denoted as CIN2+ in this paper) require further diagnosis and treatment. 
Thus, it is very important to distinguish CIN2+ from CIN1/Normal for early detection of cervical dysplasia. 

Among cervical cancer screening tests, digital cervicography is a low-cost and easy-to-access option that is suitable for low-resource regions in the world.  Images acquired by digital cervicography are called cervigrams and they can be analyzed for CIN detection and classification. 
 
While previous works on cervical cancer detection mostly rely on supervised methods~\cite{jusman2014intelligent,kim2013data,song2015multimodal,xu2017multi}, large datasets annotated by experts are required. However, labeling such data is expensive and error-prone. To mitigate this issue, we focus on semi-supervised learning (SSL) algorithms, which use a small number of labeled data while exploiting a large pool of unlabeled data to improve the model performance.
More specifically, we propose a semi-supervised approach based on graph embedding and visual features extracted with convolutional neural networks for cervical dysplasia classification. We start with visual features extracted by a pre-trained convolutional neural network. A novel graph convolutional network (GCN) model augmented by a feature encoder is developed. The GCN, with each node representing one image, enables us to effectively leverage the inter-image similarity even for unlabeled instances. Unlike previous works where the adjacency matrix in a GCN is fixed, e.g.~\cite{kipf2016semi}, we propose to use an encoder to transform visual features to an embedding space where the feature similarities are calculated. Thus, our GCN is equipped with a feature encoder to update the adjacency matrix during learning. Experiments using a varying number of labeled samples show that our semi-supervised model outperforms other baselines in all metrics significantly, especially when the number of labeled samples used is very small ($7.25\%$ labeled). We further perform an ablation study to validate the importance of our proposed learnable GCN.
\vspace{-5pt}
\section{Related Work}
\vspace{-5pt}
\label{sec:rw}
\subsection{Cervical Dysplasia Classification}
In existing literature~\cite{jusman2014intelligent,kim2013data,song2015multimodal}, various supervised learning methods have been used for cervical dysplasia classification, including neural networks, support vector machines (SVM), k-Nearest Neighbors (KNN), linear discriminant analysis (LDA), and decision trees. 
Xu~\emph{et al.}~\cite{xu2017multi} investigated the feasibility of developing an image-based automated screening method for early detection of cervical cancer. They explored different supervised learning methods on various types of features extracted from Cervigrams. 
Zhang~\emph{et al.}~\cite{zhang2010discriminative} proposed a discriminative sparse representation for tissue classification in Cervigrams. 
Lee~\emph{et al.}~\cite{lee1991integration} developed a system which integrated multiple classifiers for cytology screening.

\subsection{Semi-Supervised Learning (SSL)}
Many graph-based approaches for semi-supervised learning have been proposed, where graph embedding learning is one of the main branches. DeepWalk~\cite{perozzi2014deepwalk} learns embeddings via the prediction of the local neighborhood of nodes, sampled from random walks on the graph. Planetoid~\cite{yang2016revisiting} retains DeepWalk’s idea of predicting proximal nodes in random walks while also injecting label information.

Recent attempts at SSL have been made with graph convolutional networks (GCNs) ~\cite{kipf2016semi} in medical image analysis. Pariso~\emph{et al.}~\cite{parisot2018disease} presented a generic framework that exploited GCNs to leverage both imaging and non-imaging information for brain analysis. 
Very recently, Kazi~\emph{et al.}~\cite{kazi2019inceptiongcn} introduced InceptionGCN, a novel architecture that captures the local and global context of heterogeneous graph structures with multiple kernel sizes. 
Compared with other SSL methods, graphs provide a powerful and intuitive way of modeling samples (as nodes) and the associations or similarities between them (as edges). By making use of the quantitative relationship between every two nodes, computable for both labeled and unlabeled samples, GCNs can perform semi-supervised node classification tasks.

\vspace{-5pt}
\section{Methodology}
\begin{figure*}
\begin{center}
\includegraphics[width=0.99\textwidth]{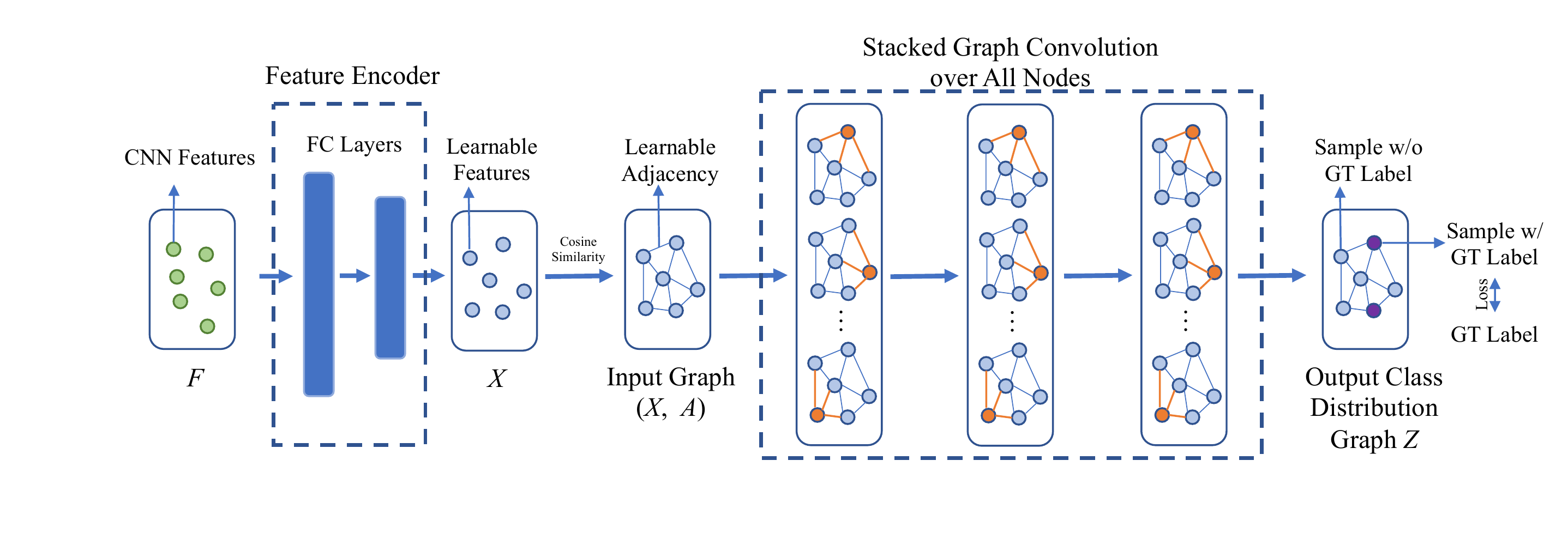}
\end{center}
\vspace{-12mm}
   \caption{Overview of our model. We use an encoder to transform the 512-D CNN extracted features $F$ to the 128-D features $X$ and build the adjacency matrix $A$ using $X$. GCN takes in $X$ and $A$ to output the class distribution $Z$ for each sample. Losses are computed over samples with available ground truth labels, and used to train both the encoder and the GCN.}
\label{fig:GCN}
\end{figure*}

\vspace{-5pt}
Our goal is to construct a semi-supervised learning pipeline for cervical dysplasia classification. To achieve this, we first fine-tune a pre-trained ResNet-18~\cite{he2016deep} model on the Cervigram dataset~\cite{xu2017multi}, and extract features for both labeled and unlabeled images using the fine-tuned CNN. We then model these visual features as nodes and their similarities as edges in a graph. Finally, we apply a graph convolutional network (GCN) with a learnable similarity metric to this graph and output the classification score for each image. More details are demonstrated in Fig.~\ref{fig:GCN}.
\vspace{-5pt}
\subsection{Feature Extraction}
\label{features}

\begin{figure}[t]
\begin{center}
\vspace{-6mm}
\includegraphics[width=0.98\linewidth]{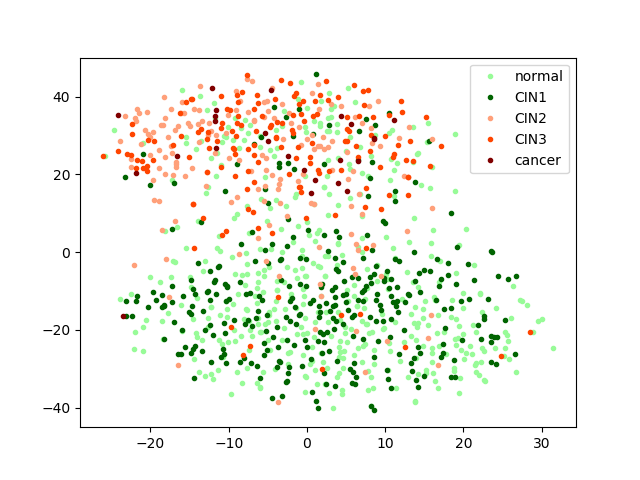}
\end{center}
\vspace{-5mm}
   \caption{t-SNE visualization of the pre-trained CNN features.}
\label{fig:features}
\vspace{-5mm}
\end{figure}

Razavian~\emph{et al.}~\cite{sharif2014cnn} demonstrated that features obtained from deep learning with a CNN are competitive in most visual recognition tasks. In this work, we investigate the performance of CNN features for cervical disease classification. We use only labeled data to fine-tune the ResNet-18 model (pretrained on ImageNet) by supervised learning on the classification task using Cervigrams. Considering the size of the dataset is relatively small, we use ResNet-18 as the backbone feature extractor in all ablation study, while extracted features are the same for different classification networks to provide a fair comparison. We extract 512-dimensional features from the last Conv layer as the CNN features. The t-SNE~\cite{maaten2008visualizing} visualization in Fig.~\ref{fig:features} illustrates that the CNN features form two distinguishable clusters representing positive (CIN2+) and negative (CIN1/Normal), respectively.

The key idea of our proposed semi-supervised learning method is to place features of both labeled and unlabeled data in a graph and leverage their correlations by building an adjacency matrix based on the similarities between features. With one node representing one image, we convert the image classification problem into a graph based node classification problem.
Different from previous GCN methods~\cite{kipf2016semi,kazi2019inceptiongcn}, to learn a better adjacency matrix, we employ a feature encoder to transform the original features to a new embedding space before computing the similarities between nodes. 
In this way, rather than being pre-determined by the original features, the adjacency matrix can be learned end-to-end through semi-supervised learning and the GCN model is more flexible. 

\subsection{ Graph Learning Architecture}
\noindent\textbf{Graph Convolutional Networks (GCNs).} 
GCN~\cite{kipf2016semi} is a variant of multi-layer convolutional neural networks that operates directly on graphs. It views the instance space as a graph where each instance is a node in the graph and the similarity between two instances is a weighted edge. Formally, consider a graph $G=(V,E)$, where
$V$ and $E$ are the sets of nodes and edges, respectively. Let $X\in R^{N\times M}$ be a matrix containing all $N$ nodes with their features, where $M$ is the dimension of the feature vectors. Let $A\in R^{N\times N}$ be the adjacency matrix of the graph and $D$ be the diagonal node degree matrix with $D_{ii} = \sum_j A_{ij}$. We assume every node is connected to itself, making $\tilde{A} = A + I_N$, where $I_N$ is the identity matrix. We follow the renormalization trick in~\cite{kipf2016semi} and denote the normalized adjacency matrix as $\hat{A} = \tilde{D}^{-\frac{1}{2}}\tilde{A}\tilde{D}^{-\frac{1}{2}}$, where $\tilde{D}_{ii} = \sum_j \tilde{A}_{ij}$. The propagation rule of the GCN layer is:
\begin{small}
\begin{equation}
    H^{(l)} = f_l(H^{(l-1)}, A) = \sigma(\hat{A}H^{(l-1)}W^{(l)}), ~~l <= L\enspace.
\end{equation}
\end{small}
The hidden features of current GCN layer $H^{(l)}$ are computed from the features of previous layer $H^{(l-1)}$ and the adjacency matrix $A$. $W^{(l)}$ are learnable parameters of the current layer. $L=3$ is the total number of GCN layers.  $H^{0} = X$ is the feature learned by the encoder, and $\sigma$ is the activation function.
The final output of our GCN model is the classification score for each node:
\begin{small}
\begin{equation}
    Z = \text{softmax}(H^{(L)}),
\end{equation}
\end{small}
We use labeled data points to calculate the cross entropy loss and train the whole network.

\vspace{3pt}
\noindent\textbf{Learnable adjacency matrix.} 
We use the cosine similarity to create the adjacency matrix $A$, since that experimental results show that cosine similarity performs the best among multiple choices:
\begin{small}
\begin{equation}
    A_{ij} = \text{Sim}(X_i, X_j) = \frac{X_i\cdot X_j}{\parallel X_i\parallel \parallel X_j \parallel}\enspace .
\end{equation}
\end{small}
The adjacency matrix $A$ is analogous to the image convolution kernel, so it would be desirable to make $A$ learnable. To achieve this, we first rewrite $A$ in the matrix form for more efficient computation:
\begin{small}
\begin{equation}
    A = \frac{XX^T}{\eta(X)\eta(X)^T}\enspace,
\end{equation}
\end{small}
where $\eta: \mathbb{R}^{N\times N} \rightarrow \mathbb{R}^{N\times 1}$ computes the L2-norm of each feature vector in $X$. Next, instead of directly assigning visual features $F$ to $X$, we use an encoder $g$ to transform $F$ to $X$ as: $X = g(F)$. 
As shown in Fig.~\ref{fig:GCN}, we model the feature encoder $g$ as a multilayer perceptron (MLP). We now convert the adjacency matrix $A$ into a function of the parameters of $g$, whose gradient can be calculated by backpropagating through $A$ to make the $A$ learnable and flexible.

\vspace{3pt}
\noindent\textbf{Loss Function.} We optimize the optimal weight $W$ by minimizing the following loss function:
\begin{small}
\begin{equation}
    \mathcal{L}=-\sum_{l \in \mathcal{Y}_{L}} \sum_{k=1}^{2} Y_{l k} \ln Z_{l k}\enspace,
\end{equation}
\end{small}
where $\mathcal{Y}_{L}$ is the set of node indices that have labels. 
Ideally, a node should only depend on a few similar nodes, so we add a sparsity-encouraging term to the loss function:  
\begin{small}
\begin{equation}
    \mathcal{L}=-\sum_{l \in \mathcal{Y}_{L}} \sum_{k=1}^{2} Y_{l k} \ln Z_{l k}+ \gamma\|A\|_{F}^{2}\enspace, 
    \label{Eq:loss 2}
\end{equation}
\end{small}
where $\gamma$ is a hyperparameter controlling the sparsity of learned graph $A$. We will discuss the effect of different loss functions in section~\ref{sec:res}.

\vspace{-7pt}
\section{Experiments}
\vspace{-5pt}
\subsection{Experiment Setup}
\noindent\textbf{Dataset.} We evaluate our method using the cervigrams dataset introduced in~\cite{xu2017multi}. The cervigrams are from a large data archive collected by the National Cancer Institute (NCI) in the Guanacaste project~\cite{herrero1997design}. The archive includes data from 10,000 anonymized women. 
The cervicography test produces two Cervigram images for a patient during her visit and the images are later sent to an expert for interpretation. Since the two images belonging to the same patient are visually similar and correlated, we randomly choose one image for each patient per visit.
In our experiments, we use 690 patient samples including 345 positive (CIN2+) and 345 negative (CIN1/Normal) images, where the negative samples are randomly chosen from 767 original samples.

\vspace{3pt}
\noindent\textbf{Baselines and Metrics.} We compare our methods against three fully supervised baselines: Support vector machines (SVM), Random Forest (RF), and ResNet-18~\cite{he2016deep}; and two graph-based semi-supervised baselines: Planetoid~\cite{yang2016revisiting} and ICA~\cite{lu2003link}.
We use common evaluation metrics, including areas under ROC curves (AUC), accuracy, sensitivity and specificity, for cervical dysplasia classification to provide a quantitative comparison. 

\vspace{3pt}
\noindent\textbf{Implementation Details.} We use 10-fold cross-validation to evaluate the classification results by different methods. The cross-validation experiment is conducted ten times with different random splits of the data. The average results are reported.
We use the open source PyTorch implementation of ResNet-18~\cite{he2016deep} to extract 512-dimensional visual features $F$ from the ROI of the Cervigrams. When pretraining the CNN, we use Adam~\cite{kingma2014adam} optimizer with the learning rate $1e-5$ and train the CNN for 80 epochs. The ROI of each training image is resized to $256\times256$ pixels and then center-cropped to $224\times224$. For the encoder $g$ of our GCN, we use an MLP with two FC layers $(256, 128)$ and tanh activation to transform CNN features to the input features $X$. For the three stacked convolutional layers in our GCN, we use ReLU activation and have $128, 128,$ and $2$ channels respectively. To prevent overfitting, we also add a dropout layer with $0.5$ dropout rate to each convolutional layer. We use Adam to optimize the GCN with learning rate $1e-4$ and weight decay $5e-5$.

\begin{table}[ht]
\caption{Classification performance comparisons between different models. All models use the same pre-trained ResNet-18 features with 50 ground truth labels (7.25\%).}\label{tab1}
\begin{center}
\vspace{-1mm}
\resizebox{0.99\columnwidth}{!}
{
\begin{tabular}{|l|c|c|c|c|}
\hline
Method &  AUC(\%) & Acc(\%) & Sensi(\%) & Speci(\%)\\
\hline
CNN\cite{he2016deep}  &  74.44 &  67.81  & 67.64 & 68.50 \\
SVM  &  74.58 &  69.81  & 70.64 & 68.95 \\
KNN &  72.11 &  66.38  & 66.68 & 66.28  \\
RF &  78.76 &  69.29  & 70.28 & 68.37  \\
Planetoid\cite{yang2016revisiting} & 80.03 & 71.74 & 73.51 & 70.83 \\
ICA\cite{lu2003link} & 69.15 & 69.86 & \textbf{86.22} & 52.09 \\
 \textbf{Ours} &  \textbf{80.66} &  \textbf{76.96} & 79.87 &  \textbf{74.14}  \\
\hline
\end{tabular}
}
\end{center}
\vspace{-8mm}
\end{table}

\subsection{Results}
\label{sec:res}
To evaluate the performance of the proposed method using a very small number of labeled examples, we only use 50 (7.25\%) labeled samples and mask the remaining samples' labels to compare our method with other baselines. As shown in Table~\ref{tab1}, our method outperforms all baselines in most metrics.

\begin{figure}[t]
\begin{center}
\includegraphics[width=0.49\textwidth]{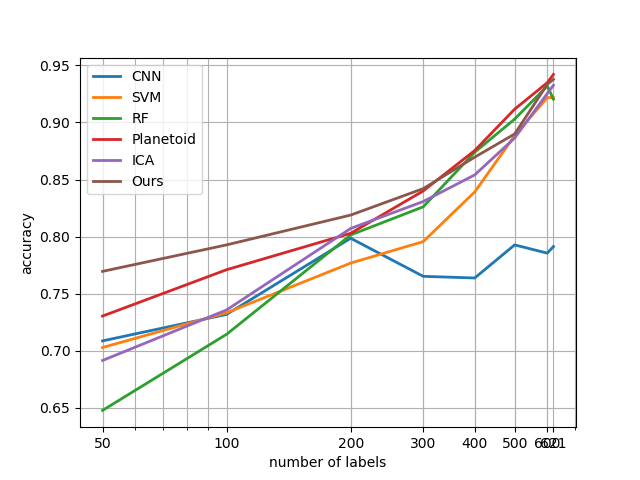}
\end{center}
   \caption{We evaluate all methods with CNN features using different numbers of labeled samples (from 50/690 to 621/690) based on accuracy.}
\label{fig:res}
\vspace{-9mm}
\end{figure}

To further assess how well our method utilizes unlabeled data, we vary the number of labeled samples from 50 to 600, and evaluate model performance at each number of labels. One can observe from Fig.~\ref{fig:res} that our method shows significant improvement over other methods when the number of labels is small, and can still achieve competitive performance when the number of labels is large. Quantitative results are provided in Table~\ref{tab2}. Note that the fully supervised learning result where we train with all the available labels (621 for training, leaving 69 for testing) achieves 93.77\% accuracy rate and 97.43\% AUC score. This performance beats that in \cite{xu2017multi}, which reported AUC and accuracy of 82.31\% and 78.41\% respectively.

\begin{table}[ht]
\caption{AUC and accuracy of our method using various numbers of ground truth labels during training.}\label{tab2}
\begin{center}
\resizebox{0.99\columnwidth}{!}
{
\begin{tabular}{|l|c|c|c|c|c|c|c|c|c|}
\hline
\# labels &  50 & 100 & 200 & 300 & 400 & 500 & 600 & 621(full) \\
\hline
 AUC(\%) & 80.66 & 84.78 & 88.25 & 91.08 & 93.12 & 95.77 & 97.07 & 97.43 \\
 Acc(\%) & 76.96 & 79.28 & 81.88 & 84.20 & 86.96 & 88.99 & 93.33 & 93.77 \\
\hline
\end{tabular}
}
\end{center}
\vspace{-5mm}
\end{table}

We also perform an ablation study to gain insight into contributions of individual components. To validate the importance of our proposed learnable adjacency matrix, we compare our model against the original GCN~\cite{kipf2016semi} without the encoder $g$ and learnable adjacency matrix. Results in Table~\ref{tab3} show that our proposed model achieves better performance in all metrics comparing with the original GCN without learnable adjacency matrix. We also measure the effect of matrix norm in Eq.~\ref{Eq:loss 2}, showing that sparsity-encouraging term brings slight improvement in AUC and sensitivity, while slightly decreasing accuracy and specificity.

\begin{table}[ht]
\caption{Ablation study with 50 groundtruth labels.}\label{tab3}
\begin{center}
\resizebox{0.99\columnwidth}{!}
{
\begin{tabular}{|l|c|c|c|c|}
\hline
Method &  AUC(\%) & Acc(\%) & Sensi(\%) & Speci(\%)\\
\hline
GCN baseline &  79.55 &  75.51  &  75.78 & \textbf{74.91}  \\
Ours w/ matrix norm &  \textbf{83.14} &   75.68  & \textbf{89.72} & 61.27 \\
Our full model &  80.66 &  \textbf{76.96} & 79.87 &  74.14  \\
\hline
\end{tabular}
}
\end{center}
\vspace{-10mm}
\end{table}
\vspace{-8pt}
\section{Conclusion}
In this paper, we have proposed a semi-supervised GCN model with learnable features and adjacency matrix for cervical dysplasia classification problem. By representing each Cervigram image with its learned feature vector and constructing a relationship graph, our proposed GCN model can infer the label of unannotated images by utilizing the quantitative relationship between them and those labeled images. Extensive experimental results demonstrate that our GCN model outperforms all baseline models with CNN features, especially when the number of utilized annotations is very small. Our proposed GCN model is general and can be easily applied to solve other medical image classification problems with very limited amount of labeled data.

\bibliographystyle{IEEEbib}
\bibliography{strings,refs}

\begin{thebibliography}{10}

\bibitem{berman2018cervical}
Nancy~R Berman and Rebecca Koeniger-Donohue,
\newblock ``Cervical cancer,''
\newblock {\em Advanced Health Assessment of Women: Clinical Skills and
  Procedures}, p. 431, 2018.

\bibitem{jusman2014intelligent}
Yessi Jusman, Siew~Cheok Ng, Abu Osman, and Noor Azuan,
\newblock ``Intelligent screening systems for cervical cancer,''
\newblock {\em The Scientific World Journal}, vol. 2014, 2014.

\bibitem{kim2013data}
Edward Kim and Xiaolei Huang,
\newblock ``A data driven approach to cervigram image analysis and
  classification,''
\newblock in {\em Color Medical Image analysis}, pp. 1--13. Springer, 2013.

\bibitem{song2015multimodal}
Dezhao Song, Edward Kim, Xiaolei Huang, Joseph Patruno, H{\'e}ctor
  Mu{\~n}oz-Avila, Jeff Heflin, L~Rodney Long, and Sameer Antani,
\newblock ``Multimodal entity coreference for cervical dysplasia diagnosis,''
\newblock {\em IEEE transactions on medical imaging}, vol. 34, no. 1, pp.
  229--245, 2015.

\bibitem{xu2017multi}
Tao Xu, Han Zhang, Cheng Xin, Edward Kim, L~Rodney Long, Zhiyun Xue, Sameer
  Antani, and Xiaolei Huang,
\newblock ``Multi-feature based benchmark for cervical dysplasia classification
  evaluation,''
\newblock {\em Pattern recognition}, vol. 63, pp. 468--475, 2017.

\bibitem{kipf2016semi}
Thomas~N Kipf and Max Welling,
\newblock ``Semi-supervised classification with graph convolutional networks,''
\newblock {\em arXiv preprint arXiv:1609.02907}, 2016.

\bibitem{zhang2010discriminative}
Shaoting Zhang, Junzhou Huang, Dimitris Metaxas, Wei Wang, and Xiaolei Huang,
\newblock ``Discriminative sparse representations for cervigram image
  segmentation,''
\newblock in {\em 2010 IEEE International Symposium on Biomedical Imaging: From
  Nano to Macro}. IEEE, 2010, pp. 133--136.

\bibitem{lee1991integration}
JS-J Lee, J-N Hwang, Daniel~T Davis, and Alan~C Nelson,
\newblock ``Integration of neural networks and decision tree classifiers for
  automated cytology screening,''
\newblock in {\em IJCNN-91-Seattle International Joint Conference on Neural
  Networks}. IEEE, 1991, vol.~1, pp. 257--262.

\bibitem{perozzi2014deepwalk}
Bryan Perozzi, Rami Al-Rfou, and Steven Skiena,
\newblock ``Deepwalk: Online learning of social representations,''
\newblock in {\em Proceedings of the 20th ACM SIGKDD international conference
  on Knowledge discovery and data mining}. ACM, 2014, pp. 701--710.

\bibitem{yang2016revisiting}
Zhilin Yang, William~W Cohen, and Ruslan Salakhutdinov,
\newblock ``Revisiting semi-supervised learning with graph embeddings,''
\newblock {\em arXiv preprint arXiv:1603.08861}, 2016.

\bibitem{parisot2018disease}
Sarah Parisot, Sofia~Ira Ktena, Enzo Ferrante, Matthew Lee, Ricardo Guerrero,
  Ben Glocker, and Daniel Rueckert,
\newblock ``Disease prediction using graph convolutional networks: Application
  to autism spectrum disorder and alzheimer’s disease,''
\newblock {\em Medical image analysis}, vol. 48, pp. 117--130, 2018.

\bibitem{kazi2019inceptiongcn}
Anees Kazi, Hendrik Burwinkel, Gerome Vivar, Karsten Kortuem, Seyed-Ahmad
  Ahmadi, Shadi Albarqouni, Nassir Navab, et~al.,
\newblock ``Inceptiongcn: Receptive field aware graph convolutional network for
  disease prediction,''
\newblock {\em arXiv preprint arXiv:1903.04233}, 2019.

\bibitem{he2016deep}
Kaiming He, Xiangyu Zhang, Shaoqing Ren, and Jian Sun,
\newblock ``Deep residual learning for image recognition,''
\newblock in {\em Proceedings of the IEEE conference on computer vision and
  pattern recognition}, 2016, pp. 770--778.

\bibitem{sharif2014cnn}
Ali Sharif~Razavian, Hossein Azizpour, Josephine Sullivan, and Stefan Carlsson,
\newblock ``Cnn features off-the-shelf: an astounding baseline for
  recognition,''
\newblock in {\em Proceedings of the IEEE conference on computer vision and
  pattern recognition workshops}, 2014, pp. 806--813.

\bibitem{maaten2008visualizing}
Laurens van~der Maaten and Geoffrey Hinton,
\newblock ``Visualizing data using t-sne,''
\newblock {\em Journal of machine learning research}, vol. 9, no. Nov, pp.
  2579--2605, 2008.

\bibitem{herrero1997design}
Rolando Herrero, Mark~H Schiffman, Concepci{\'o}n Bratti, Allan Hildesheim,
  Ileana Balmaceda, Mark~E Sherman, Mitchell Greenberg, Fernando C{\'a}rdenas,
  V{\'\i}ctor G{\'o}mez, Kay Helgesen, et~al.,
\newblock ``Design and methods of a population-based natural history study of
  cervical neoplasia in a rural province of costa rica: the guanacaste
  project,''
\newblock {\em Revista Panamericana de Salud P{\'u}blica}, vol. 1, pp.
  362--375, 1997.

\bibitem{lu2003link}
Qing Lu and Lise Getoor,
\newblock ``Link-based classification,''
\newblock in {\em Proceedings of the 20th International Conference on Machine
  Learning (ICML-03)}, 2003, pp. 496--503.

\bibitem{kingma2014adam}
Diederik~P Kingma and Jimmy Ba,
\newblock ``Adam: A method for stochastic optimization,''
\newblock {\em arXiv preprint arXiv:1412.6980}, 2014.

\end{thebibliography}

\end{document}